\documentclass[sigconf]{acmart}
\usepackage{booktabs}

\AtBeginDocument{%
  \providecommand\BibTeX{{%
    \normalfont B\kern-0.5em{\scshape i\kern-0.25em b}\kern-0.8em\TeX}}}

 \acmConference[nn]{nn}{nn}{nn}

\begin{document}

\title[WaLDORf]{WaLDORf: Wasteless Language-model Distillation On Reading-comprehension}

\author{James Yi Tian}
\email{james.tian@sap.com}

\affiliation{%
  \institution{SAP Labs}
}

\author{Alexander P. Kreuzer}
\email{alexander.kreuzer@sap.com}

\affiliation{%
  \institution{SAP Labs}
}

\author{Pai-Hung Chen}
\email{pai-hung.chen@sap.com}

\affiliation{%
  \institution{SAP Labs}
}

\author{Hans-Martin Will}
\email{hans-martin.will@sap.com}

\affiliation{%
  \institution{SAP Labs}
}

\renewcommand{\shortauthors}{Tian, et al.}

\begin{abstract}
Transformer based Very Large Language Models (VLLMs) like BERT, XLNet and RoBERTa, have recently shown tremendous performance on a large variety of Natural Language Understanding (NLU) tasks. However, due to their size, these VLLMs are extremely resource intensive and cumbersome to deploy at production time. Several recent publications have looked into various ways to distil knowledge from a transformer based VLLM (most commonly BERT-Base) into a smaller model which can run much faster at inference time. Here, we propose a novel set of techniques which together produce a task-specific hybrid convolutional and transformer model, WaLDORf, that achieves state-of-the-art inference speed while still being more accurate than previous distilled models.
\end{abstract}




\maketitle

\section{Introduction}
The recent emergence of VLLMs\cite{peters2018deep, devlin2018bert, yang2019xlnet, liu2019roberta, radford2019language, shoeybi2019megatron, raffel2019exploring} have made a wide variety of previously extremely difficult NLU tasks feasible. From an academic perspective, the leader boards of a broad set of NLU benchmarks, such as the General Language Understanding Evaluation (GLUE\cite{wang2018glue} or Super-GLUE\cite{wang2019superglue}) and the Stanford Question Answering Dataset (SQuAD\cite{rajpurkar2016squad, rajpurkar2018know}), are dominated by fine-tuned versions of such VLLMs, generally based on transformers\cite{vaswani2017attention} and the attention mechanism\cite{bahdanau2014neural, luong2015effective, britz2017massive, graves2014neural, xu2015show, cheng2016long}. From industry, real world problems are now being solved by putting these VLLMs into production. Recently, both Google and Microsoft have announced that they use some (possibly distilled) version of a VLLM in their search algorithms. In academia, a plethora of research has arisen not only towards advancing and augmenting these state-of-the-art VLLMs\cite{peters2019knowledge, conneau2019cross, lample2019large, sun2019ernie, dai2019transformer, joshi2019spanbert} but also towards studying their inner workings\cite{wallace2019universal, wallace2019allennlp, si2019does, wallace2019nlp, tenney2019bert, michel2019sixteen, clark2019does, arkhangelskaia2019whatcha, hewitt2019structural, niven2019probing} - a field sometimes called "BERTology".  

Looking over the entire NLU landscape, it has become increasingly apparent that deploying a (fine-tuned) VLLM into production can generate a lot of value. However the resource requirements of deploying a VLLM into production can be prohibitively expensive. VLLMs, being very large, require both a lot of memory and a lot of computation for inference. Therefore, there has been much recent work into distilling knowledge from a VLLM into a much smaller, faster model\cite{sun2019patient, jiao2019tinybert, tang2019distilling, sanh2019distilbert}. Our work continues that line of work by examining several novel techniques which, together with previously published techniques, produce a leaner, faster but still highly accurate distilled model. This distilled model, in turn, can produce massive cost-savings for production environments. In concrete terms, while maintaining Exact Match (EM) and F1 scores on SQuAD v2.0 higher than TinyBERT, 4-layer Patient Knowledge Distillation of BERT (BERT-PKD), and 4-layer DistilBERT, we have achieved 1.24x, 3.1x, and 3.1x speed up from those models respectively (4-layer BERT-PKD and 4-layer DistilBERT have the same model architecture and so inference speed is the same). Compared with BERT-Base and BERT-Large, we achieved 9.13x and 28.63x speed up respectively.  

For this work, we are looking into \textit{task specific} distillation for the SQuAD v2.0 task. I.e. we do not perform model distillation on the pre-training phase of BERT. We chose to look more deeply into task specific distillation so that we could explore task specific data augmentation and its effects on model distillation in more detail. We chose specifically the SQuAD v2.0 task because it is relatively difficult and flexible. Recent advances in NLU have proposed to frame a wide variety of NLU tasks into the Question-Answering format\cite{raffel2019exploring, mccann2018natural}.

The major contributions (which we will describe in detail in section \ref{Methodology}) of our work are:
\begin{itemize}
\item With inspiration from convolutional auto encoders, we use convolutions, max pooling, and up sampling to shrink the sequence length of inputs to the transformer encoder blocks and then expand the outputs of those encoder blocks back to the original sequence length. (Section \ref{architecture})  
\item We modify the hidden-state distillation and attention-weight distillation by using average pooling and max pooling respectively to align the sequence dimension of our model with the teacher model. (Section \ref{training}) 
\item We modify the patient knowledge distillation procedure to build layers up from the bottom one layer at a time to reduce the covariate shift experienced by each encoder block. (Section \ref{training})
\item We fine-tune a pre-trained sequence-to-sequence VLLM to produce high quality questions from given contexts to perform task specific data augmentation. (Section \ref{dataAugment}). 
\end{itemize}   

The rest of our paper is organized as follows: in section \ref{Related} we discuss some related work and models from which we drew inspiration; in section \ref{Methodology} we describe in detail our methodology, including model architecture, model distillation procedure, and data augmentation; in section \ref{Results} we present all of our results, including results from our ablation study; in section \ref{Discussion} we discuss our findings in detail; and finally in section \ref{Conclusion} we present a few concluding remarks and directions for future work. We also present some details to reproduce our results in the appendix, section \ref{appendix}.

\section{Related Work} \label{Related}
Model compression has been explored in the past in several parallel paths. There is a whole field of study devoted to weight quantization, weight pruning, and weight encoding to reduce the size of large neural networks - often achieving impressive results\cite{han2015deep, han2015learning, lee2018retraining, zhao2019improving, lin2016fixed, cheng2017survey, lecun1990optimal, frankle2018lottery, liu2018rethinking, zhou2019deconstructing}. Recently, there have been efforts to apply weight quantization and pruning to transformer models as well\cite{voita2019analyzing, prato2019fully, cheong2019transformers}. For this work, we instead focus on the parallel path(s) of exploration into compressing large transformers using architectural changes or model distillation.

The recently released ALBERT\cite{lan2019albert} approached model compression from the point of view of making architectural and training objective changes to the transformer model. The authors of ALBERT use the following three main techniques: 1) Factorized embedding parameterization to greatly reduce the number of weights in the embedding layer, 2) Weight sharing among all encoder blocks of the transformer, greatly reducing the number of weights in the encoder stack, and 3) Using an inter-sentence coherence loss instead of a next sentence prediction loss during BERT-like pre-training. With these architectural and training modifications, ALBERT produces impressive results and achieved state-of-the-art accuracy, for their xx-large model, on a wide variety of NLU tasks. ALBERT-base also achieves quite high accuracy while reducing the number of parameters to 12 million (compared with 110 million for BERT-Base and 335 million for BERT-Large). However, because data still must be processed through every layer of ALBERT, even though the weights are shared between all the layers, the inference speed improvement achieved by ALBERT is not so striking. ALBERT-Base achieves 21.1x speed up from an extra large version of BERT built by the authors of ALBERT (BERT-XLarge) which is itself 17.7x \textit{slower} than BERT-Base. This means that ALBERT-Base achieves only roughly 1.2x speed up from BERT-Base itself. For our purposes, we wanted to look at much faster architectures. Therefore, we only incorporated the factorized embedding parameterization idea from ALBERT. Other possibly faster architectures have of course also been explored for language modeling\cite{merity2019single}. 

Model distillation was first applied to distilling ensemble model knowledge into a single model\cite{bucilua2006model} and then expanded to other domains\cite{hinton2015distilling}. In this vein, we point out three recent efforts at model distillation of BERT: DistilBERT\cite{sanh2019distilbert}, BERT-PKD\cite{sun2019patient}, and TinyBERT\cite{jiao2019tinybert}. DistilBERT took inspiration from model distillation as put forth by Hinton et. al. and distilled knowledge from the softmax outputs produced by BERT-Base into a 6 encoder block (also called 6-layer) version of BERT. DistilBERT has half the number of encoder blocks as BERT-Base but is otherwise identical to BERT-Base (in terms of hidden dimension size, the number of attention heads, etc.). BERT-PKD goes further and distils the hidden state activations, layer-by-layer, from BERT-Base (and in one experiment, BERT-Large) into 3 and 6 encoder block versions of BERT-Base. The authors of BERT-PKD show that patient knowledge distillation generally outperforms last-layer distillation. Finally, TinyBERT is a 4 encoder block version of BERT-Base that goes even further and includes distillation objectives for the attention scores and the embedding layer. In addition, TinyBERT shrinks the hidden dimension sizes of BERT-Base to produce a much smaller and much faster model than either DistilBERT or BERT-PKD. The accuracy on SQuAD v2.0 attained by TinyBERT also outperforms those attained by DistilBERT and BERT-PKD, for a given number of encoder blocks, so for our work we generally benchmark against TinyBERT. We take inspiration from TinyBERT and perform a version of their layer-by-layer patient knowledge distillation. However, different from TinyBERT, we also included some architectural changes that necessitated changes to the knowledge distillation procedure. Furthermore, we did not reproduce the language modeling pre-training used by BERT and all these previously mentioned distillations. Instead, we relied purely on data augmentation to achieve state-of-the-art results.   


\section{Methodology} \label{Methodology}
\subsection{Problem Statement}
For this work, we are working specifically in the framework of question-answering as posed by the SQuAD v2.0 dataset. To perform question-answering, we are given a question and a context and we must find the concrete answer to the question. More specifically, SQuAD v2.0\cite{rajpurkar2018know} is an extractive, text-based, question-answering dataset which tasks us to find spans of text within the context which answer the given question or, if the question is unanswerable given the context, to return a null result. Formally, given a tuple $(Q, C)$, where $Q$ is the question, and $C$ is the context consisting of a passage of text, find $A$, which is a span in $C$, that answers the question $Q$ or return a null result if $Q$ is not answerable given $C$. For training, we are given tuples of training examples $(Q,C,A)$, while for evaluation, we are given the tuple $(Q,C)$ and a set of ground truth answers $\{A\}$ that are variants of the answer for the given question.  

We use two primary metrics to evaluate our model for accuracy, the EM and F1 metrics. The EM score divides the number of examples for which our model's answer, $\hat{A}$, exactly matched \textit{any} of the ground truth answers in $\{A\}$ into the total number of evaluation examples. The F1 score averages the maximum overlap between $\hat{A}$ and $\{A\}$ over all the evaluation examples. These are the same two metrics used by the SQuAD v2.0 dataset itself.

\subsection{Model Architecture}\label{architecture}
\begin{figure*}[!htb]
  \center{\includegraphics[width=\textwidth]
  {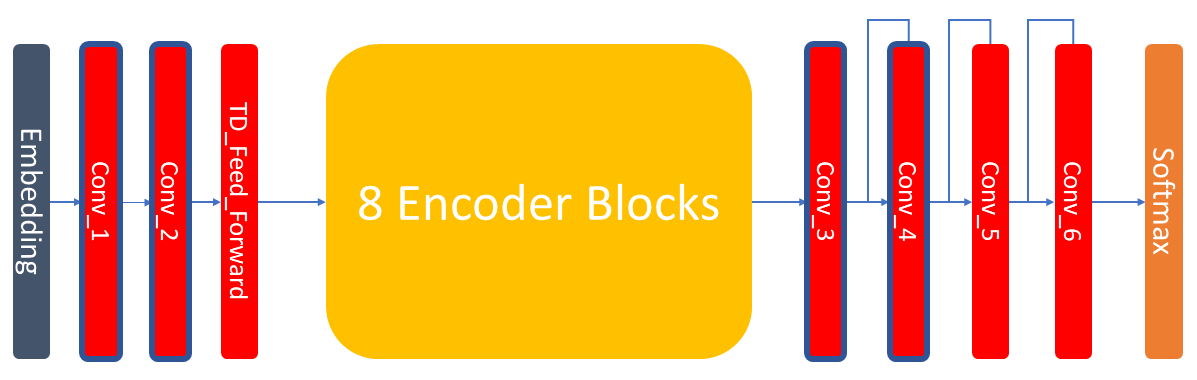}}
  \caption{\label{fig:architecture} A visualization of WaLDORf's architecture. There are 8 transformer encoder blocks sandwiched between 1D convolutional layers along the sequence length direction. All convolutional layers use padding to preserve sequence length. Conv\_1 and Conv\_2 layers are paired with maxpooling to reduce sequence length by a total factor of 4. Conv\_3 and Conv\_4 layers are paired with up sampling to restore the sequence length to its original size. Skip connections are used after up sampling layers to skip past convolutional layers but before layer normalization.}
\end{figure*}

WaLDORf is a hybrid model that incorporates convolutional layers into the transformer encoder stack architecture of BERT\cite{devlin2018bert}; see figure \ref{fig:architecture} for an illustration. Taking inspiration from ALBERT, we chose the embedding dimension to be smaller than the internal hidden dimension used in the encoder blocks\cite{lan2019albert}. We use the same word piece tokenization scheme as BERT (which is itself a sub word tokenization scheme similar to Byte Pair Encoding\cite{gage1994new}) in order to allow distillation of the embedding layer as described in section \ref{training}. Before feeding data into the 8 encoder blocks, we use a time-distributed feed-forward layer to expand the dimensions of the hidden representation to equal that of the encoder blocks. The major architectural insight we bring forth with WaLDORf is to use convolutional layers to shrink and then re-expand the input sequence dimension. The self-attention mechanism employed by the transformer encoder blocks have complexity which is quadratic in the input sequence length\cite{vaswani2017attention}. By using two 1D convolutions (along the sequence length dimension) coupled with two max-pooling layers, we shrink the input sequence lengths to the encoder blocks by a factor of 4. In turn, this means each self-attention mechanism needs to do only 1/16th the amount of work. We then use two more 1D convolutions (since there are no transposed convolutions in 1 dimension) coupled with two up sampling layers to re-expand the sequence length dimension back to its original length. By greatly reducing the amount of work carried out by the self-attention mechanism, we are able to make the encoder blocks run much faster so that we can have a deeper model without sacrificing speed. The trade-off is, of course, that the model must learn to encode the input information in a "smeared out" way; with roughly every 4 word-pieces getting 1 internal representation. This trade-off may affect different tasks differently depending on how sensitive a certain task is to specific word-pieces. However, we found that even for the SQuAD v2.0 task, which intuitively should be relatively sensitive to specific words or word-pieces, since the task is specifically about finding spans within the text, we were still able to achieve good accuracy with our model.

The 8 encoder blocks used in our model are the standard transformer encoder blocks used by BERT which have been scaled down to increase inference speed. A detailed list of architecture hyperparmeters can be seen in table \ref{table:archHyper}. The self-attention mechanisms in our model use 16 heads of attention like BERT-Large instead of 12 heads of attention like BERT-Base because we chose to use BERT-Large as our teacher model. In total, our model has $24.6~\textrm{M}$ parameters which is a bit less than a quarter of the number of parameters of BERT-Base.       

\begin{table}
\caption{hyperparameters for WaLDORf's architecture}
\centering
\begin{tabular}{lr}
\toprule
\multicolumn{2}{c}{Architecture Hyperparameters} \\
\midrule
\# Encoder Blocks & $8$ \\
\# Encoder Convolutional Layers & $2$ \\
\# Decoder Convolutional Layers & $4$ \\
$conv\_1$ filters & $96$ \\
$conv\_2$ filters & $192$ \\
$conv\_3-6$ filters & $480$ \\
Embedding Size & $96$ \\
Hidden Size & $480$ \\
Feed-Forward Size & $1440$ \\
\# Attention Heads & $16$ \\
\# Total Parameters & $24.6~\textrm{M}$ \\
\bottomrule
\label{table:archHyper}
\end{tabular}
\end{table}

\subsection{Training Objectives and Procedure} \label{training}
To train our model, we used a modified version of the patient knowledge distillation procedure. We have 5 main training losses: 
\begin{itemize}
\item The distillation loss over the embedding layer: $\mathcal{L}_e$.
\item The distillation loss over the hidden state representations inside the 8 encoder blocks: $\mathcal{L}_h$.
\item The distillation loss over the attention scores used to calculate the attention weights inside the 8 self-attention mechanisms: $\mathcal{L}_a$. 
\item The distillation loss over the final softmax layer which corresponds to the vanilla model distillation loss introduced by Hinton et. al.: $\mathcal{L}_d$.
\item The ground truth loss (for data for which we have the ground truth): $\mathcal{L}_g$.  
\end{itemize}

The student model (our model) learns from a teacher model which we chose to be BERT-Large Whole-Word-Masking (BERT-Large WWM) which is a variant of BERT-Large that was trained using a masked token objective where the masked tokens were ensured to be over whole words instead of word pieces as in the original. The first 4 losses mentioned, $\mathcal{L}_e,~~\mathcal{L}_h,~~\mathcal{L}_a,~~\mathcal{L}_d $, are all losses for the student model with respect to the teacher model. The final loss $\mathcal{L}_g$ is the loss with respect to the ground truth and would be the only loss available to us if we did not perform model distillation. As we will show in section \ref{ablation}, if we use only $\mathcal{L}_g$   to try to train our model, our results are extremely poor.

The embedding loss, $\mathcal{L}_e$, is a simple mean squared error loss that measures how different the embedding representations of the student model are from the teacher model. Given $E^s_i\in \mathbb{R}^{l\times e'}$, where $E^s_i$ are the student model embeddings for the $i~\textrm{th}$ text sequence, $l$ is the input sequence length, and $e'$ is the embedding size of the student model, and $E^t_i\in \mathbb{R}^{l\times e}$, where $E^t_i$ are the teacher model embeddings for the $i~\textrm{th}$ text sequence, and $e$ is the teacher model embedding size, then:
\begin{equation}
\mathcal{L}_e = \frac{1}{B} \sum_{i=0}^B \textrm{MSE}(E^t_i, E^s_i W^E)
\end{equation}
Where MSE is the mean-squared error, $B$ is the batch size, and $W^E\in\mathbb{R}^{e'\times e}$ is a learned tensor that projects the student embeddings to the same dimension size as the teacher embeddings.

Both the hidden loss, $\mathcal{L}_h$, and the attention scores loss $\mathcal{L}_a$ are mean squared error losses that tries to match the hidden states and attention weights within the 8 encoder blocks of the student model to 8 encoder blocks teacher model. We choose to follow the PKD-skip scheme\cite{sun2019patient} and use every 3$^{rd}$ encoder block in the teacher model for distillation. Just like for the embedding loss, we also have to project the student model hidden states to the same dimension size as the teacher model's. In addition, we also use an average pooling over the teacher model's hidden states along the sequence length dimension to shrink the sequence length of the teacher model to match the sequence length inside the student model's encoder blocks. Because WaLDORf used convolutions to shrink the sequence length used in the encoder blocks, we can't match hidden states, word for word, between the two models. If we think of hidden states analogously to word vectors, then we are essentially asking WaLDORf to approximate the average of 4 word vectors with each of its hidden states. This procedure is perhaps the simplest way to make the dimensions of our student model align with the targets given by the teacher model. Empirically, this procedure gives us good results. Thus, given $H^s_{j,i}\in\mathbb{R}^{l/4\times d'}$, where $H^s_{j,i}$ are the hidden state outputs for the $i~\textrm{th}$ text sequence of the $j~\textrm{th}$ encoder block of the student model, and $d'$ is the hidden dimension size of the student model, and $H^t_{j,i}\in\mathbb{R}^{l\times d}$, where $H^t_{j,i}$ are the hidden state outputs for the $i~\textrm{th}$ text sequence of the $j~\textrm{th}$ encoder block of the teacher model, and $d$ is the hidden dimension size of the teacher model, then:
\begin{equation}
\mathcal{L}_{h,j} = \frac{1}{B}\sum_{i=0}^B \textrm{MSE}(\textrm{AVG}(H^t_{3j+2,i}),H^s_{j,i}W^h)
\end{equation} 
Where AVG denotes a 1-D average pooling with size 4 and stride 4 along the sequence length dimension:
\begin{equation}
\textrm{AVG}(H_{j,i})_{a,b} = \underset{k_x\in [0,4)}{\textrm{avg}} ~H_{j,i,a+k_x,b}~
\end{equation}
$\mathcal{L}_{h,j}$ is the hidden loss of the $j~\textrm{th}\in \{0,...,7\}$ encoder block, and $W^h\in\mathbb{R}^{d'\times d}$ is a learned matrix that projects student model hidden dimension sizes to those of the teacher model. Note that we use the same $W^h$ for every encoder block because we want the distilled learning to be learned by WaLDORf itself and not by these projection matrices which will be discarded at inference time. 

For the attention score loss, instead of using average pooling to make sequence dimensions match, we switch to using max pooling. We decided to use max pooling here because we want the student model to learn the \textit{strongest} attention scores rather than an average of attention scores. Hence, given $\xi^s_{j,i}\in\mathbb{R}^{l/4\times l/4}$, where $\xi^s_{j,i}$ are the attention scores for the $i~\textrm{th}$ text sequence of the $j~\textrm{th}$ encoder block, and $\xi^t_{j,i}\in\mathbb{R}^{l\times l}$, where $\xi^t_{j,i}$ are the attention scores for the $i~\textrm{th}$ text sequence of the $j~\textrm{th}$ encoder block, then:
\begin{equation}
\mathcal{L}_{a,j} = \frac{1}{B}\sum_{i=0}^B \textrm{MSE}(\textrm{MAX}(\xi^t_{3j+2,i}),\xi^s_{j,i}) 
\end{equation} 
Where MAX denotes a 2-D max pooling operation with size $4\times 4$ and stride $4\times 4$:
\begin{equation}
\textrm{MAX}(\xi_{j,i})_{a,b} = \underset{k_x\in [0,4),k_y\in [0,4)}{\textrm{max}} ~\xi_{j,i,a+k_x,b+k_y}
\end{equation}
and $\mathcal{L}_{a,j}$ is the attention score loss of the $j~\textrm{th}\in \{0,...,7\}$ encoder block. Note that here we are operating on attention \textit{scores} and not the attention weights. The attention scores have not had a softmax applied to them and so do not necessarily add up to 1, hence we are free to perform a simple max pooling without destroying any normalization since the softmax will be applied after. One detail that should be noted is that proper masking needs to be maintained on the attention score loss so that the system does not try to learn minor changes to the masking value.   

The distillation loss over the final softmax layer follows closely the distillation loss introduced in Hinton et. al.\cite{hinton2015distilling}:
\begin{equation}
\begin{split}
\mathcal{L}_d = \frac{T^2}{B}\sum_{i=0}^B\textrm{softmax}(z^t_i/T) \\ 
\cdot ~\textrm{log\_softmax}(z^s_i/T)
\end{split}
\end{equation}
Where $T\in [1,\infty]$ is a temperature hyperparameter and $z^t_i\in\mathbb{R}^l,~z^s_i\in\mathbb{R}^l$ are teacher and student logits for the $i~\textrm{th}$ text sequence respectively. We multiply the loss by the factor of $T^2$ to ensure that the gradients are of the correct scale. 

Lastly, the ground truth loss is the standard cross entropy loss with the ground truth labels:
\begin{equation}
\mathcal{L}_g = \frac{1}{B}\sum_{i=0}^B y_i\textrm{log\_softmax}(z^s_i)
\end{equation}
Where $y_i$ is the ground truth label for the $i~\textrm{th}$ text sequence. This ground truth loss is only available for data for which we have the ground truth labels. When we perform data augmentation, as discussed in the next section, we will no longer have access to ground truth labels and so this loss will not apply. Note that for both the ground truth loss and the distillation loss over the final softmax, there are two sets of logits (and labels), one for the start position and one for the end position. The softmax functions are taken over the sequence length dimension.

\begin{figure}[!htb]
  \center{\includegraphics[width=\columnwidth]{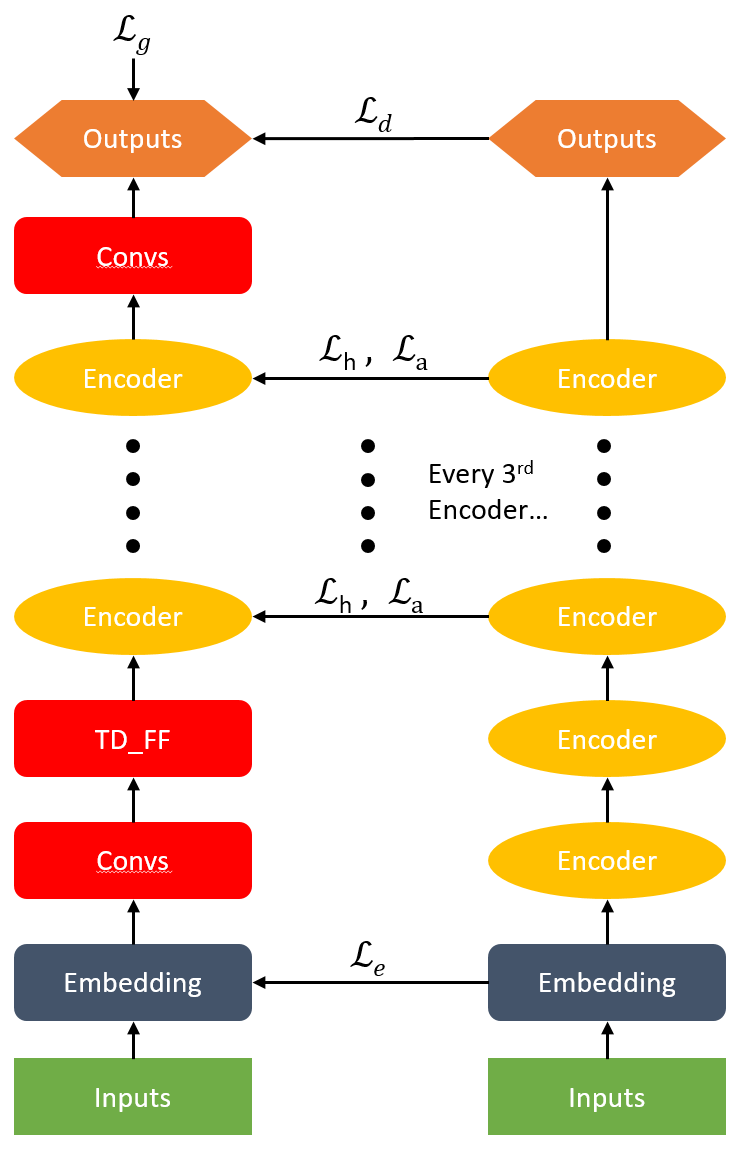}}
  \caption{\label{fig:pkd} The patient knowledge distillation procedure used for WaLDORf. The left-side model is the student and the right-side model is the teacher. Attention scores and hidden states are distilled from every 3rd teacher encoder block.}
\end{figure}

Figure \ref{fig:pkd} shows our version of the patient knowledge distillation procedure in detail. 

To train our model, we found it was very beneficial to train each layer from the bottom up (or left to right in figure \ref{fig:architecture}) so that we start with training only using the embedding loss, then combine that loss with the hidden state and attention score losses for layer 0, then for layer 1, etc. After each intermediate layer is trained for some time, we then add the final softmax layer distillation loss and the ground truth loss. This training procedure makes sense intuitively because it should reduce the internal covariate shift experienced by each layer during training (the same goal as for batch normalization\cite{ioffe2015batch}). Since we have targets available for each layer in the student model, with the exception of the convolutional layers, we can use those intermediate targets to ensure that the inputs to each layer have stabilized somewhat by the time that that layer is trained. Thus, our final loss function is:
\begin{equation}
\begin{split}
\mathcal{L}(\tau) = \alpha\mathcal{L}_e + \sum_{j=0}^{7}\chi_j(\tau)(\beta\mathcal{L}_{h,j}+\gamma\mathcal{L}_{a,j}) \\
+\chi_8(\tau)(\delta\mathcal{L}_d+\epsilon\mathcal{L}_g)
\end{split}
\end{equation}
Where $\alpha,\beta,\gamma,\delta,\epsilon$ are real number hyperparameters and: 
\begin{equation}
\chi_j(\tau)=\Theta(\tau-(j+1)\tau^*)
\end{equation}
Where $\Theta$ is the Heaviside step function, $\tau$ is the current global training step and $\tau^*$ is an integer hyperparameter specifying how many training steps to take for each layer. We will show in our ablation study in section \ref{ablation} that this layer-by-layer building up of the loss function significantly improves accuracy.  

\subsection{Data Augmentation} \label{dataAugment}
A large part of why VLLMs perform so well on downstream tasks is due to the massive amount of data that they pretrain on\cite{yang2019xlnet, radford2019language, raffel2019exploring}. For distilled architectures, such as DistilBERT or BERT-PKD, which keep hidden dimensions all the same size as a teacher model (e.g. BERT-Base) and just use fewer encoder blocks, the student encoder blocks can be initialized with the weights from a subset of encoders blocks of a trained teacher model. For distilled architectures, such as TinyBERT or our model, which change the hidden dimension sizes as well as the number of encoder blocks, there is no obvious way to initialize the weights of the model with weights from the teacher model. Hence, distilled architectures also often reproduce the pretraining phase of BERT before being fine-tuned (with or without additional distillation) on downstream tasks. Here, we are focused on \textit{task-specific} distillation and so we do not desire to reproduce the pretraining, language modeling, phase. However, not reproducing the pretraining phase would put our model at a huge disadvantage simply due to the much smaller coverage of input data that our model would be trained on. Hence, to help balance out the imbalance in data volume, we chose to perform extensive data augmentation. It should be noted though that even with data augmentation, the final volume of data, as measured by total token count, that we train on is still much smaller than the pretraining data used to train BERT.

To perform data augmentation, we sourced 500,000 additional paragraphs, from 280,250 English Wikipedia articles which did not correspond to any paragraphs or articles found in the SQuAD v2.0 dataset (training or dev). For each paragraph, we automatically generated a question using a fine-tuned Text-to-Text Transfer Transformer, T5-Large\cite{raffel2019exploring}. We fine-tuned T5-Large $T5:C\rightarrow Q$ by using question-context pairs $(Q,C)$ from the SQuAD v2.0 training dataset itself. Our empirical experiments will show (in section \ref{Results}) that this data augmentation increased  the end-to-end accuracy by a large margin. For comparison, the SQuAD v2.0 training dataset has roughly 130,000 questions on 20,000 paragraphs sourced from 500 articles.


\section{Results} \label{Results}
Overall, WaLDORf achieved significant speed up over previous state-of-the-art distilled architectures while maintaining a higher level of accuracy on SQuAD v2.0.
 
\subsection{Inference Speed}
\begin{table*}[t]
\caption{Results obtained from testing for inference speed. $avg\_time$ is the time to perform inference once over the SQuAD v2.0 cross validation set averaged over many trials. $rel\_speedup$ is relative speed up as compared with BERT-Base. See section \ref{speedTest} for details on how the testing was conducted.}
\centering
\begin{tabular}{lrrrrrr}
\toprule
\multicolumn{2}{l}{Inference Speed Results} \\
\midrule
Model & $n\_encoders$ & $hidden\_size$  & $int\_dim$ & $n\_params$ & $avg\_time$ & $rel\_speedup$ \\ 
\midrule
BERT-Base & $12$ & $768$ & $3072$ & $109.5~\textrm{M}$ & $106.2~\textrm{s}$ & 1x \\
BERT-Large & $24$ & $1024$ & $4096$ & $335.1~\textrm{M}$ & $333.0~\textrm{s}$ & 0.3x \\
\midrule
BERT-6 & $6$ & $768$ & $3072$ & $67.0~\textrm{M}$ & $54.2~\textrm{s}$ & 2x \\
BERT-4 & $4$ & $768$ & $3072$ & $52.8~\textrm{M}$ & $36.0~\textrm{s}$ & 2.9x \\
BERT-2 & $2$ & $768$ & $3072$ & $38.6~\textrm{M}$ & $18.4~\textrm{s}$ & 5.8x \\
TinyBERT & $4$ & $312$ & $1200$ & $14.4~\textrm{M}$ & $14.4~\textrm{s}$ & 7.4x \\
WaLDORf & $8$ & $480$ & $1440$ & $24.6~\textrm{M}$ & $\textbf{11.6}~\textbf{s}$ & \textbf{9.1x} \\
\bottomrule
\label{table:speedTest}
\end{tabular}
\end{table*}

Detailed results of our inference speed testing are presented in table \ref{table:speedTest}. As clearly shown, our model is the fastest model to perform inference over the SQuAD v2.0 cross validation set by a wide margin. Because of our use of convolutional layers, we were able to have a model which is deeper and has larger hidden dimensions while still being faster than TinyBERT. DistilBERT and BERT-PKD would both fall under the BERT-6 umbrella since in terms of architecture both are simply 6 encoder block versions of BERT-Base. BERT-PKD also has a 3 encoder block version, but we can see that our model is much faster than even a 2 encoder block version of BERT-Base. 

\subsection{Accuracy Performance}
\begin{table}[htb]
\caption{WaLDORf's accuracy performance on SQuAD v2.0 as compared with TinyBERT, BERT-PKD, and DistilBERT. BERT-PKD, DistilBERT and TinyBERT results are taken from TinyBERT paper. TinyBERT-6 has scaled up internal dimension sizes to match those of BERT-Base. BERT-Base and BERT-Large-WWM results are from models that we fine-tuned. We used BERT-Large-WWM as our teacher model.}
\centering
\begin{tabular}{lrr}
\toprule
\multicolumn{2}{l}{SQuAD v2.0 Results} \\
\midrule
Model & EM & F1  \\ 
\midrule
BERT-Base & $71.8$ & $75.2$ \\
BERT-Large-WWM & $82.6$ & $85.6$\\
\midrule
BERT-PKD-4 &  $60.8$ & $64.6$ \\
DistilBERT-4 & $60.6$ & $64.1$ \\
TinyBERT & $65.3$ & $68.6$ \\
WaLDORf & $\textbf{66.0}$ & $\textbf{70.3}$ \\
\midrule
BERT-PKD-6 & $66.3$ & $69.8$ \\
DistilBERT-6 & $66.0$ & $69.5$ \\
TinyBERT-6 & $69.9$ & $73.4$ \\
\bottomrule
\label{table:accuracy}
\end{tabular}
\end{table}

Our model's performance on SQuAD v2.0 is benchmarked against other models in table \ref{table:accuracy}. The BERT-PKD-4 and DistilBERT-4 models presented have internal dimensions identical with BERT-4 from table \ref{table:speedTest}. Thus, our model would be 3.1x faster than those models for inference while being significantly more accurate. WaLDORf is also somewhat more accurate than TinyBERT ($+ 0.7$ EM, $+ 1.7$ F1) while still being 1.24x faster. To obtain EM and F1 scores higher than ours, one would have to go to 6-layer versions of those models for which our model would be 4.5x faster. TinyBERT-6 has internal dimension sizes scaled to be the same as BERT-Base as well and so would also be 4.5x slower than WaLDORf.  

\subsection{Ablation Study} \label{ablation}
\begin{table}[htb]
\caption{Ablation study for WaLDORf. The baseline model is trained on just ground truth labels. "Softmax Distil" is, in addition, trained on the softmax probabilities provided by the teacher model. "All Layer Distil" adds on the embedding layer, hidden state, and attention score losses as described in section \ref{training}. "Slow Build" builds each layer of the student model one-by-one. "Data Augment", as described in \ref{dataAugment}, was the final technique we added and gave us our final results. +EM and +F1 denote improvements over the previous best model. }
\centering
\begin{tabular}{lrrrr}
\toprule
\multicolumn{2}{l}{Ablation Study} \\
\midrule
 & EM & F1 & +EM & +F1  \\ 
\midrule
Baseline & $34.0$ & $37.6$ & & \\
\midrule
$~$+ Softmax Distil & $42.9$ & $45.8$ &  $8.9$ & $8.2$\\
$~$+ All Layer Distil &  $45.9$ & $49.7$ & $3.0$ & $3.9$ \\
$~$+ Slow Build & $57.7$ & $62.2$ & $\textbf{11.8}$ & $\textbf{12.5}$ \\
\midrule
$~$+ Data Augment & $\textbf{66.0}$ & $\textbf{70.3}$ & $8.3$ & $8.1$ \\
\bottomrule
\label{table:ablation}
\end{tabular}
\end{table}

Results from our ablation study are provided in table \ref{table:ablation}. Each additional technique shown in the table was added on to the sum of the previously used techniques. As we can clearly see, each technique which we added produced significant improvements. With the addition of each new technique, we also explored a whole set of hyperparameters. Results are presented for the best set of hyperparameters found within a given framework. When all the techniques are combined, we get a very significant absolute improvement of $32.0$ in EM score and $32.7$ in F1 score over the baseline.

The biggest incremental improvement in accuracy that we saw was in moving from trying to distill every layer all at once to slowly building up the layers from the bottom embedding layer up. This is likely due to the fact that training every layer all at once introduces a significant covariate shift for every layer's inputs during training. Stabilizing the inputs to each layer by training layers from a bottom up fashion appears to make optimization much easier.  
 

\section{Discussion} \label{Discussion}
\subsection{Convolutions and Autoencoding}
The core architectural change that we bring forth in this work is the use of convolutions, max pooling, and up sampling to reduce and then re-expand the sequence length dimension. We introduced this architectural change with the hope that, analogous to convolutional autoencoders, WaLDORf will find an efficient way to compress the information contained within an example into a shorter sequence. By attaining accuracy performance higher than previous state-of-the-art, we have shown that it is in fact feasible for the model to compress information in this way. Information can be effectively encoded by the input convolutions, processed by the encoder blocks, and then decoded by the output convolutions.  In addition, we also showed that by using max pooling and average pooling we are able to use only the signals given by the "important" attention weights and the averaged hidden states from the teacher model to make our student model perform strongly. These properties may be somewhat task dependent, but even for the SQuAD v2.0 task we were able to attain a high level of accuracy. 

Due to the quadratic dependence of the complexity of the self-attention mechanism on the sequence length, our architecture will actually see larger relative gains in speed the longer the input sequence length is. We chose a sequence length of 384 for our tests because that is the sequence length used by BERT for the SQuAD data set. Other transformer-based VLLMs may use even longer sequence lengths in order to deal better with longer paragraphs, e.g. XLNet uses a max sequence length of 512 for SQuAD. For circumstances where the max input sequence length is even longer than 384, we expect to gain even larger relative speed ups.

\subsection{Distillation}
Model distillation is currently a very active area of research. Having access to a teacher model opens up the possibility of using a huge variety of techniques to improve a student model's performance. The distillation techniques available often depend on the architecture of the student model and how similar it is to the teacher model. A model, like BERT-PKD or DistilBERT, that is essentially a copy of a teacher model with just fewer layers can straight-forwardly distil a subset of the layers in the teacher model. If one scales down the internal dimensions of the model, like in the case of TinyBERT, then one needs to introduce some way to project the student model's hidden dimensions to those of the teacher model or vice versa. That projection operator introduces new degrees of freedom to the problem and the distillation procedure is no longer as straight forward. In our case, we had to also modify the sequence-length dimension of the teacher model's signals in order to perform intermediate-layer distillation. 

In the case that the student model is entirely different from the teacher model, e.g. when a LSTM-based student model tries to learn from a transformer based teacher model as in Tang et. al.\cite{tang2019distilling}, then perhaps the only easily distilled layer is the final layer and no intermediate layer distillation is feasible. However, we (and others) have shown that there is significant performance improvements to be had by distilling also the intermediate layers and operations (e.g. attention scores) of a VLLM. It is uncertain, at this point, how much of the performance improvement is due to the difference in the raw information contained within the training signals (softmax signals vs intermediate layers and operations signals) and how much is due to a difference in how much those signals help the student model to optimize. The concrete difference in these scenarios would be that if the latter is true, then student models with architectures wildly different from a teacher model may still attain a high level of accuracy if the last layer distillation is carried out with a very strong optimization scheme. If the former scenario is true, then we may expect that there is no method by which we can optimize a student model using only the signals from the last layer as strongly as we could by using signals from all the layers.       
 
\subsection{Data Augmentation}
As we showed in section \ref{ablation}, data augmentation gives us a significant 8-point boost in EM and F1 scores over distillation using just the SQuAD v2.0 dataset itself. This result shows that even with roughly 130,000 question-context-answer tuples, SQuAD v2.0 is still not large enough of a dataset to fully train our student model using pure model distillation. The set of targets that the student model had to hit, given only SQuAD v2.0 inputs, was already enormously large. The embedding layer targets are of size $E \in \mathbb{R}^{n\times l\times e} $, hidden layer targets are of size $H\in\mathbb{R}^{8\times n\times (l/4) \times d}$ and the attention score targets are of size $\xi\in \mathbb{R}^{8\times n\times (l/4) \times (l/4)}$, where $n$ denotes the total number of examples. For a BERT-Large-WWM teacher model over the SQuAD v2.0 dataset, these tensors would comprise roughly 1.2 Terabytes of data. The sheer size of this data makes for a very rich set of signals given to the student model, however, given the success of our data augmentation experiments, the input space provided by SQuAD v2.0 itself did not appear rich enough for the student model to fully exploit those signals.

A huge benefit of performing model distillation and student-teacher learning is the ease with which "labeled" data can be obtained from unlabeled data. Data augmentation for us was simplified further by the introduction of the powerful T5 model which we fine-tuned to automatically generate questions given a passage. But, in the absence of such a model, generating a huge amount of unlabeled data in the context of a production environment is often quite feasible. In the case of question-answering, for example, we may be able to mine user-generated questions and run those through a teacher model to provide data for the student model to train on. For task specific knowledge distillation, we have shown that purely performing data augmentation is a viable alternative to reproducing the language modeling pre-training phase.

\section{Conclusion and Further Work} \label{Conclusion}
In this work, we have shown a set of techniques that together successfully advance the state-of-the-art in accuracy and speed of distilled language models for the SQuAD v2.0 task. There are quite many promising directions for future work within this field. Of course, the main considerations would be in pushing the limits of model distillation itself. Perhaps by cleverly using adapter layers of some kind, it will become possible in the future to distil the knowledge contained within intermediate layers of a VLLM to a much smaller model that has a totally different architecture from the teacher model. Or maybe a combination of model distillation techniques and model pruning and weight sharing could bring even greater speed gains without loss in accuracy. Lastly, beyond the practical benefits, model distillation techniques may be used to probe just how much the size of a VLLM is necessary to performing specific NLU tasks. Model distillation is an extremely rich area of research well deserving additional exploration.

\begin{acks} \label{Acknowledgements}
The authors would like to thank all the members of the SAP Innovation Center Newport Beach as well as our collaborators at the University of California, Irvine, for helpful discussions and inspiration.
\end{acks}

\bibliographystyle{ACM-Reference-Format}
\bibliography{bibliography}

\appendix

\section{Methodology Details}\label{appendix}

\subsection{Inference Speed Testing} \label{speedTest}
To perform inference speed testing, we set up a Nvidia V100 GPU running Tensorflow 1.14 and CUDA 10. We must point out here that inference speed and relative speed up are quite sensitive to the exact testing conditions. For example, changing the input sequence length or prediction batch size may change both the absolute inference speed and relative speed up one architecture gets over another. Inference speed and relative speed up are therefore task dependent and are not universal. These reasons are why the relative speed up reported by us may not be entirely consistent with the relative speed up reported by others. In particular, while TinyBERT found 9.4x speed up over BERT-Base for their particular testing environment, which used a batch size of 128 on the QNLI training set and a maximum sequence length of 128 on a NVidia K80 GPU, within our testing environment, TinyBERT showed only 7.4x speedup over BERT-Base. For our test environment, we chose to use a maximum sequence length of 384 and a batch size of 32. We ran inference multiple times on the SQuAD v2.0 dev set and averaged the run-times to average out any network or memory caching dependent performance issues.    

\subsection{Training Hyperparameters}
\begin{table}
\caption{hyperparameters for training WaLDORf}
\centering
\begin{tabular}{lr}
\toprule
\multicolumn{2}{c}{Training Hyperparameters} \\
\midrule
Init Learning Rate & $2\times 10^{-4}$ \\
Batch Size & $24$ \\
Dropout & $0$ \\
Adam Epsilon & $1\times 10^{-5}$ \\
Init Temperature & $5$ \\
Epochs & $35$ \\
$\alpha$ & $1$ \\
$\beta$ & $1.2$ \\
$\gamma$ & $1.4$ \\
$\delta$ & $1$ \\
$\epsilon$ & $1$ \\
$\tau^*$ & $57500$ \\
\bottomrule
\label{table:trainHyper}
\end{tabular}
\end{table}

We present the hyperparameters used to train WaLDORf in table \ref{table:trainHyper}. These are the hyperparameters we found gave us the best results. The initial learning rate was kept constant over the "building up" phase of training where layers and encoder blocks are slowly being added to the training objective. Once the model reached the final phase of training and all of the loss functions were in play, the learning rate was decayed linearly to 0 by the end of training. The initial temperature was also decayed linearly to 1 for the final phase of training. For our choice of data volume, batch size, and steps per layer ($t^*$), 35 epochs of training corresponds roughly to 1 million global steps with 517,500 of those steps being spent during the build-up phase of training.

\subsection{Hardware}
All model training was performed using a single Google v3 TPU running TensorFlow 1.14. Data parallelization among the 8 TPU cores was handled by the TPUEstimator API. As mentioned in subsection \ref{speedTest}, all inference speed testing was performed using a single Nvidia V100 GPU.  

\end{document}